\documentclass[10pt,twocolumn,letterpaper]{article}

\usepackage{cvpr}              

\usepackage[dvipsnames]{xcolor}

\usepackage{wrapfig}
\usepackage{multirow}
\usepackage{float}
\usepackage{multicol}
\usepackage{cuted}
\usepackage{stfloats}

\definecolor{cvprblue}{rgb}{0.21,0.49,0.74}
\usepackage[pagebackref,breaklinks,colorlinks,citecolor=cvprblue]{hyperref}

\def\paperID{*****} 
\def\confName{CVPR}
\def\confYear{2024}

\title{Sapiens: Foundation for Human Vision Models\vspace{-0.2in}} 

\author{ 
Rawal Khirodkar, Timur Bagautdinov, Julieta Martinez, Su Zhaoen, \\ Austin James, Peter Selednik, Stuart Anderson,  Shunsuke Saito \\
{\urlstyle{sf} \href{https://about.meta.com/realitylabs/codecavatars/sapiens}{https://about.meta.com/realitylabs/codecavatars/sapiens}}
}
\begin{document}

\twocolumn[{
    \vspace*{-0.7in}
    \maketitle
    \begin{figure}[H]
    \hsize=\textwidth
    \centering
    \vspace*{-0.4in}
    \includegraphics[width=0.85\textwidth, height=0.82\textwidth]{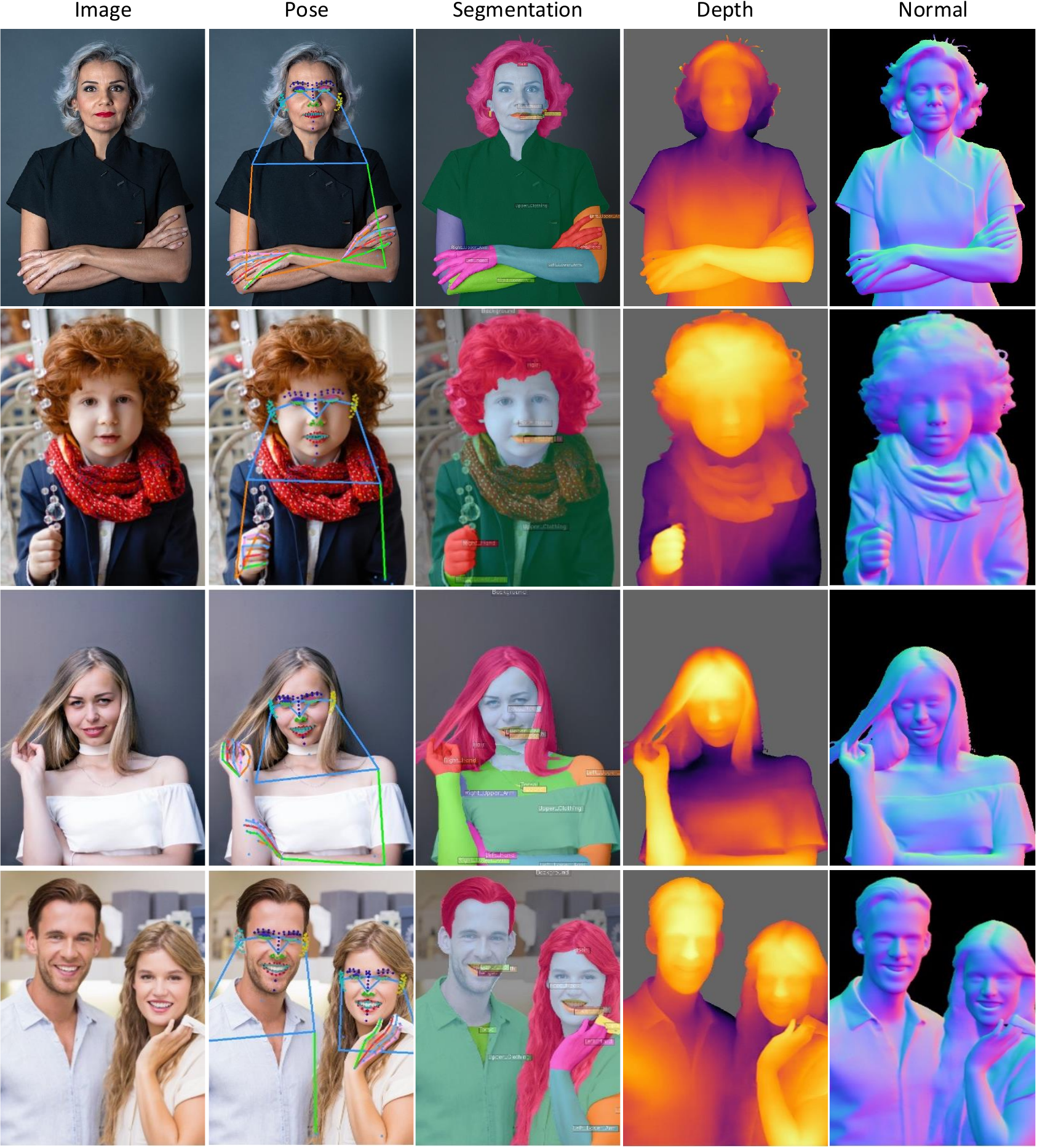}
    \vspace*{-0.1in}
    \caption{\textit{Sapiens} models are finetuned for four human tasks - 2D pose estimation, body-part segmentation, depth prediction and normal prediction. Our models generalize across a variety of in-the-wild face, upper-body, full-body and multi-person images.}
    \label{figure:introduction}
    \end{figure}
}]

\begin{abstract}
\vspace{-0.2in}
    We present \textit{Sapiens}, a family of models for four fundamental human-centric vision tasks
-- 2D pose estimation, body-part segmentation, depth estimation, and surface normal prediction. 
Our models natively support 1K high-resolution inference and are extremely easy to adapt for individual tasks 
by simply fine-tuning models pretrained on over $300$ million in-the-wild human images. We observe that, given the same computational budget, self-supervised pretraining on a curated dataset of human images significantly boosts the performance for a diverse set of human-centric tasks. The resulting models exhibit remarkable generalization to in-the-wild data, even when labeled data is scarce or entirely synthetic. Our simple model design also brings scalability -- model performance across tasks improves as we scale the number of parameters from $0.3$ to $2$ billion. Sapiens consistently surpasses existing baselines across various human-centric benchmarks. We achieve significant improvements over the prior state-of-the-art on Humans-5K (pose) 
by $7.6$ mAP, Humans-2K (part-seg) by $17.1$ mIoU, Hi4D (depth) by 22.4\% relative RMSE, and THuman2 (normal) by $53.5$\% relative angular error.
\vspace{-0.2in}
\end{abstract}

\begin{strip}
\begin{quote}
\vspace{-0.1in}
\centering
``\textit{Sapiens}---pertaining to, or resembling modern humans.''
\vspace{-0.2in}
\end{quote}
\end{strip}

\section{Introduction}
\label{sec:introduction}
Recent years have witnessed remarkable strides towards generating photorealistic humans in 2D~\cite{chan2019everybody,zhang2023adding,drobyshev2022megaportraits,hu2023animate} and 3D~\cite{lombardi2019neural,saito2020pifuhd,weng2022humannerf,xiu2022icon}. The success of these methods is greatly attributed to the robust estimation of various assets such as 2D keypoints~\cite{cao2016realtime,lin2014microsoft}, fine-grained body-part segmentation~\cite{zhang2019pose2seg}, depth~\cite{yang2024depth}, and surface normals~\cite{saito2020pifuhd,xiu2022econ}. However, robust and accurate estimation of these assets is still an active research area, and complicated systems to boost performance for individual tasks often hinder wider adoption. Moreover, obtaining accurate ground-truth annotation in-the-wild is notoriously difficult to scale. Our goal is to provide a unified framework and models to infer these assets in-the-wild to unlock a wide range of human-centric applications for everybody.

We argue that such human-centric models should satisfy three criteria: generalization, broad applicability, and high fidelity. Generalization ensures robustness to unseen conditions, enabling the model to perform consistently across varied environments. Broad applicability indicates the versatility of the model, making it suitable for a wide range of tasks with minimal modifications. High fidelity denotes the ability of the model to produce precise, high-resolution outputs, essential for faithful human generation tasks. This paper details the development of models that embody these attributes, collectively referred to as \textit{Sapiens}.

Following the insights from~\cite{oquab2023dinov2,el2024scalable,singh2023effectiveness}, leveraging large datasets and scalable model architectures is key for generalization. For broader applicability, we adopt the pretrain-then-finetune approach, enabling post-pretraining adaptation to specific tasks with minimal adjustments. This approach raises a critical question: \textit{What type of data is most effective for pretraining?} Given computational limits, should the emphasis be on collecting as many human images as possible, or is it preferable to pretrain on a less curated set to better reflect real-world variability? 
Existing methods often overlook the pretraining data distribution in the context of downstream tasks. To study the influence of pretraining data distribution on human-specific tasks, we collect the Humans-300M dataset, featuring $300$ million diverse human images. These unlabelled images are used to pretrain a family of vision transformers~\cite{dosovitskiy2020image} from scratch, with parameter counts ranging from 300M to 2B. 

Among various self-supervision methods for learning general-purpose visual features from large datasets~\cite{zhou2021ibot,el2024scalable,chen2020simple,he2020momentum,bao2021beit,he2022masked}, we choose the masked-autoencoder (MAE) approach~\cite{he2022masked} for its simplicity and efficiency in pretraining. MAE, having a single-pass inference model compared to contrastive or multi-inference strategies, allows processing a larger volume of images with the same computational resources. For higher-fidelity, in contrast to prior methods, we increase the native input resolution of our pretraining to $1024$ pixels, resulting in a ${\sim}4\times$ increase in FLOPs compared to the largest existing vision backbone~\cite{singh2023effectiveness}. 
Each model is pretrained on $1.2$ trillion tokens. Table~\ref{table:introduction} outlines a comparison with earlier approaches. For finetuning on human-centric tasks~\cite{cao2018openpose,zhang2019pose2seg,yang2024depth,wang2015designing}, we use a consistent encoder-decoder architecture. The encoder is initialized with weights from pretraining, while the decoder, a lightweight and task-specific head, is initialized randomly. Both components are then finetuned end-to-end. We focus on four key tasks - 2D pose estimation, body-part segmentation, depth, and normal estimation, as shown in Fig.~\ref{figure:introduction}.


\begin{table}[b]
\captionsetup{font=small}
\small
\vspace*{-0.1in}
\begin{center}

\resizebox{3.4in}{!}{
    \setlength{\tabcolsep}{5pt}
    \renewcommand{\arraystretch}{1.1}
    \begin{tabular}{llrrrl}
    \toprule
    \textbf{Method} & \textbf{Dataset} & \textbf{\#Params} & \textbf{GFLOPs}  & \textbf{Image size} &\textbf{Domain}  \\
    \midrule
    DINO~\cite{caron2021emerging} & ImageNet1k & 86 M & $17.6$ & 224 & General\\ 
    iBOT~\cite{zhou2021ibot} & ImageNet21k & 307 M & $61.6$ & 224 & General\\ 
    DINOv2~\cite{oquab2023dinov2} & LVD-142M & 1 B & $291.0$ & 224 & General\\ 
    ViT-6.5B~\cite{singh2023effectiveness} & IG-3B & 6.5 B & $1657.0$ & 224 & General\\ 
    AIM~\cite{el2024scalable} & DFN-2B & 6.5 B & $1657.0$ & 224 & General \\ 
    \midrule
    Sapiens (Ours) &  Humans-300M & 2 B & $8709.0$ & 1024 & Human \\
    \bottomrule
    \end{tabular}
}
\vspace{-0.1in}
\caption{Comparison of state-of-the-art pretrained vision models. Sapiens adopts a higher resolution backbone on a large dataset of in-the-wild human images.}
\label{table:introduction}
\vspace*{-0.1in}

\end{center}
\end{table}

Consistently with prior studies~\cite{zlateski2018importance,kato2018improving}, we affirm the critical impact of label quality on the model's in-the-wild performance. Public benchmarks~\cite{jin2020whole,gong2017look,couprie2013indoor} often contain noisy labels, providing inconsistent supervisory signals during 
model fine-tuning. At the same time, it is important to utilize fine-grained and precise annotations to align closely with our primary goal of 3D human digitization. To this end, we propose a substantially denser set of 2D whole body keypoints for pose estimation and a detailed class vocabulary for body part segmentation, surpassing the scope of previous datasets (please refer to Fig.~\ref{figure:introduction}). 
Specifically, we introduce a comprehensive collection of $308$ keypoints encompassing the body, hands, feet, surface, and face. Additionally, we expand the segmentation class vocabulary to $28$ classes, covering body parts such as the hair, tongue, teeth, upper/lower lip, and torso. To guarantee the quality and consistency of annotations and a high degree of automation, we utilize a multi-view capture setup to collect pose and segmentation annotations. We also utilize human-centric synthetic data for depth and normal estimation, leveraging $600$ detailed scans from RenderPeople~\cite{RenderPeople_3DPeople} to generate high-resolution depth maps and surface normals.

We show that the combination of domain-specific large-scale pretraining with limited, yet high-quality annotations leads to robust in-the-wild generalization. Overall, our method demonstrates an effective strategy for developing highly precise discriminative models capable of performing in real-world scenarios without the need for collecting a costly and diverse set of annotations.

\noindent
Our contributions are summarized as follows.
\begin{itemize}
\itemsep0em 
  \item We introduce \textit{Sapiens}, a family of vision transformers pretrained on a large-scale dataset of human images.
  \item This study shows that simple data curation and large-scale pretraining significantly boost the model's performance with the same computational budget.
  \item Our models, fine-tuned with high-quality or even synthetic labels, demonstrate in-the-wild generalization.
  \item The first 1K resolution model that natively supports high-fidelity inference for human-centric tasks, achieving state-of-the-art performance on benchmarks for 2D pose, body-part segmentation, depth, and normal estimation.
\end{itemize}

\section{Related Work}
\label{sec:related_works}
Our work explores the limits of training large architectures on a large number of in-the-wild human images. We build on prior work from different areas: pretraining at scale, human vision tasks, and large vision transformers.

\vspace{2mm}\noindent 
\textbf{Pretraining at Scale.} The remarkable success of large-scale pretraining~\cite{devlin2018bert, tay2021scale} followed by task-specific finetuning for language modeling~\cite{achiam2023gpt, touvron2023llama, touvron2023llama2,team2023gemini,jiang2023mistral, brown2020language} has established this approach as a standard practice. Similarly, computer vision methods~\cite{abnar2021exploring, goyal2021self, bai2023sequential, zhang2023gpt, rombach2022high, oquab2023dinov2, radford2021learning, saharia2022photorealistic, el2024scalable, el2021large} are progressively embracing extensive data scales for pretraining. The emergence of large datasets, such as LAION-5B~\cite{schuhmann2022laion}, Instagram-3.5B~\cite{mahajan2018exploring}, JFT-300M~\cite{sun2017revisiting}, LVD-142M~\cite{oquab2023dinov2}, Visual Genome~\cite{krishna2017visual}, and YFCC100M~\cite{thomee2016yfcc100m}, has enabled the exploration of a data corpus well beyond the scope of traditional benchmarks~\cite{russakovsky2015imagenet, lin2014microsoft, krizhevsky2009learning}. Salient work in this domain includes DINOv2~\cite{oquab2023dinov2}, MAWS~\cite{singh2023effectiveness}, and AIM~\cite{el2024scalable}. DINOv2 achieves state-of-the-art performance in generating self-supervised features by scaling the contrastive iBot~\cite{zhou2021ibot} method on the LDV-142M dataset~\cite{oquab2023dinov2}. MAWS~\cite{singh2023effectiveness} studies the scaling of masked-autoencoders (MAE)~\cite{he2022masked} on billion images. AIM~\cite{el2024scalable} explores the scalability of autoregressive visual pretraining similar to BERT~\cite{devlin2018bert} for vision transformers~\cite{dosovitskiy2020image}. In contrast to these methods which mainly focus on general image pretraining or zero-shot image classification, we take a distinctly human-centric approach: our models leverage a vast collection of human images for pretraining, subsequently fine-tuning for a range of human-related tasks.

\vspace{2mm}\noindent 
\textbf{Human Vision Tasks.}
The pursuit of large-scale 3D human digitization~\cite{levoy2000digital,bartol2021review,halstead1996reconstructing,lowe1987three} remains a pivotal goal in computer vision~\cite{bojic2022metaverse}. Significant progress has been made within controlled or studio environments~\cite{ma2021pixel,lombardi2021mixture,saito2020pifuhd,alldieck2022photorealistic,lombardi2019neural,lawrence2021project,kocabas2023hugs}, yet challenges persist in extending these methods to unconstrained environments~\cite{du2023avatars}. To address these challenges, developing versatile models capable of multiple fundamental tasks such as keypoint estimation~\cite{chen2018cascaded,fang2017rmpe,he2017mask,huang2017coarse,newell2016stacked, papandreou2017towards,sun2019deep,khirodkar2021multi,xiao2018simple}, body-part segmentation~\cite{xia2017joint,fang2018weakly,xia2016zoom,gong2018instance,gong2017look,luo2018macro,gong2018instance}, depth estimation~\cite{yang2024depth,eigen2014depth,bhat2021adabins,li2022binsformer,ranftl2020towards,bhat2023zoedepth,guizilini2023towards,jafarian2021learning}, and surface normal prediction~\cite{xiu2022econ,saito2019pifu,wang2015designing,eigen2015predicting,fouhey2013data,ladicky2014discriminatively,barron2014shape,barron2013intrinsic} from images in natural settings is crucial. In this work, we aim to develop models for these essential human vision tasks which generalize to in-the-wild settings.

\vspace{2mm}\noindent 
\textbf{Scaling Architectures.} Currently, the largest publicly-accessible language models contain upwards of 100B parameters~\cite{hoffmann2022training}, while the more commonly used language models~\cite{taori2023alpaca, touvron2023llama2} contain around 7B parameters.  In contrast, Vision Transformers (ViT)~\cite{dosovitskiy2020image}, despite sharing a similar architecture, have not been scaled to this extent successfully. While there are notable endeavors in this direction, including the development of a dense ViT-4B~\cite{chen2022pali} trained on both text and images, and the formulation of techniques for the stable training of a ViT-22B~\cite{dehghani2023scaling}, commonly utilized vision backbones still range between 300M to 600M parameters~\cite{he2016deep,fang2023eva,liu2022swin,dai2021coatnet} and are primarily pretrained at an image resolution of about $224$ pixels. 
Similarly, existing transformer-based image generation models, such as DiT~\cite{peebles2023scalable} use less than $700$M parameters, and operate on a highly compressed latent space.
To address this gap, we introduce Sapiens - a collection of large, high-resolution ViT models that are pretrained natively at a $1024$ pixel image resolution on millions of human images.

\section{Method}
\label{sec:method}
\begin{figure}
    \centering
    \includegraphics[width=0.5\linewidth]{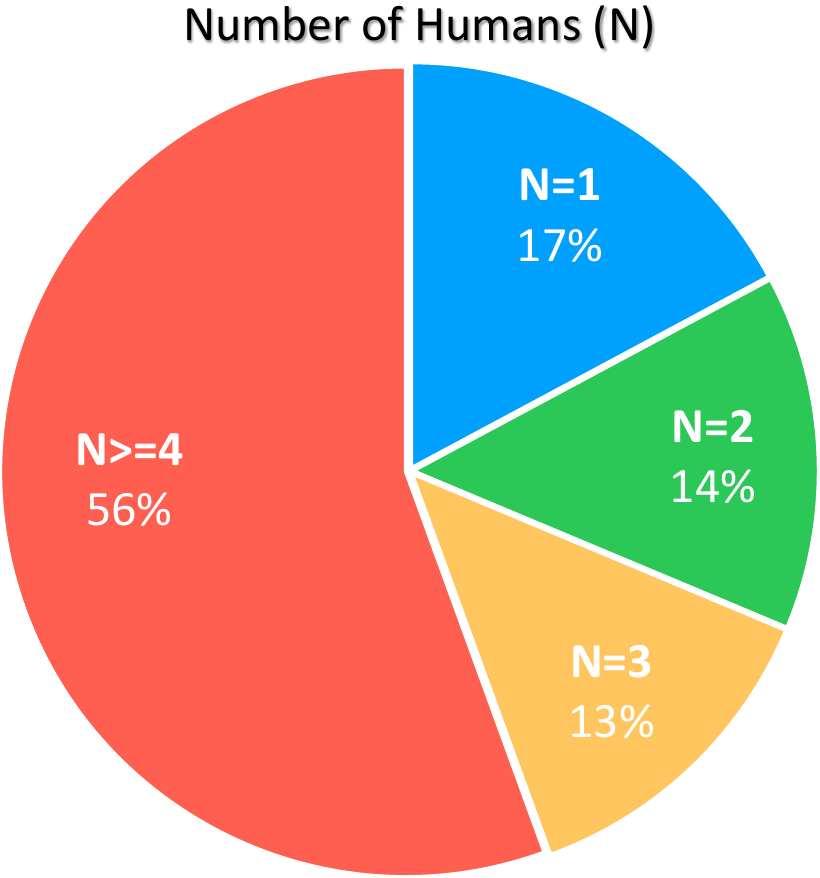}
    \caption{Overview of number of humans per image in the Humans-300M dataset.}
    \label{figure:dataset}
\vspace{-0.2in}
\end{figure}

\begin{figure*}
\centering
\includegraphics[width=0.9\linewidth]{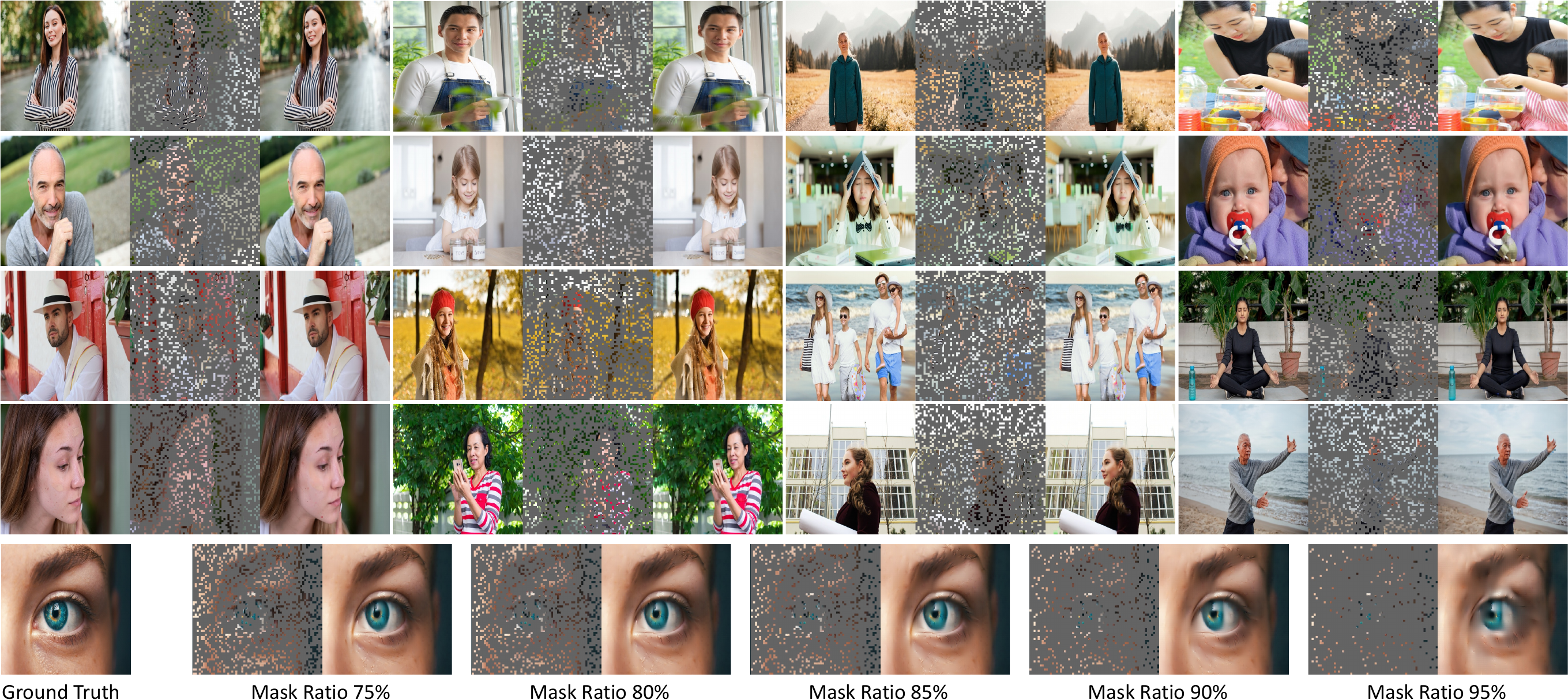}
\caption{\textit{Sapiens} reconstruction on unseen images. \textit{Top}: Each triplet contains the ground truth (left), the masked image (center), and the MAE reconstruction (right), with a masking ratio of $75\%$, a patch size of $16$, and an image size of $1024$. \textit{Bottom}: Varying the mask ratio between [0.75, 0.95] during inference reveals a minimal reduction in quality, underscoring the model's understanding of human images.}
\vspace*{-0.2in}
\label{figure:method}
\end{figure*}

\subsection{Humans-300M Dataset}
We utilize a large proprietary dataset for pretraining of approximately $1$ billion in-the-wild images, focusing exclusively on human images.
The preprocessing involves discarding images with watermarks, text, artistic depictions, or unnatural elements. Subsequently, we use an off-the-shelf person bounding-box detector~\cite{wu2019detectron2} to filter images, retaining those with a detection score above $0.9$ and bounding box dimensions exceeding $300$ pixels. Fig.~\ref{figure:dataset} provides an overview of the distribution of the number of people per image in our dataset, noting that over $248$ million images contain multiple subjects.

\subsection{Pretraining}
We follow the masked-autoencoder~\cite{he2022masked} (MAE) approach for pretraining. Our model is trained to reconstruct the original human image given its partial observation. Like all autoencoders, our model has an encoder that maps the visible image to a latent representation and a decoder that reconstructs the original image from this latent representation. Our pretraining dataset consists of both single and multi-human images; each image is resized to a fixed size with a square aspect ratio. Similar to ViT~\cite{dosovitskiy2020image}, we divide an image into regular non-overlapping patches with a fixed patch size. A subset of these patches is randomly selected and masked, leaving the rest visible. The proportion of masked patches to visible ones is defined as the masking ratio, which remains fixed throughout training. We refer to MAE~\cite{he2022masked} for more details. Fig.~\ref{figure:method} (\textit{Top}) shows the reconstruction of our pretrained model on unseen human images. 

Our models exhibit generalization across a variety of image characteristics including scales, crops, the age and ethnicity of subjects, and number of subjects. 
Each patch token in our model accounts for $0.02$\% of the image area compared to $0.4$\% in standard ViTs, a $16 \times$ reduction - this provides a fine-grained inter-token reasoning for our models. Fig.\ref{figure:method} (\textit{Bottom}) shows that even with an increased mask ratio of $95\%$, our model achieves a plausible reconstruction of human anatomy on held-out samples.

\subsection{2D Pose Estimation}
We follow the top-down paradigm, which aims to detect the locations of $K$ keypoints from an input image $\mathbf{I} \in \mathbb{R}^{H \times W \times 3}$. Most methods pose this problem as heatmap prediction, where each of $K$ heatmaps represents the probability of the corresponding keypoint being at any spatial location. Similar to~\cite{xu2022vitpose}, we define a pose estimation transformer, $\mathcal{P}$, for keypoint detection. The bounding box at training and inference is scaled to $H \times W$ and is provided as an input to $\mathcal{P}$. Let $\mathbf{y} \in \mathbb{R}^{H \times W \times K}$ denote the $K$ heatmaps corresponding to the ground truth keypoints for a given input $\mathbf{I}$. The pose estimator transforms input $\mathbf{I}$ to a set of predicted heatmaps, $\hat{\mathbf{y}} \in \mathbb{R}^{H \times W \times K}$, such that $\hat{\mathbf{y}} = \mathcal{P}(\mathbf{I})$. $\mathcal{P}$ is trained to minimize the mean squared loss $\mathcal{L}_\text{pose} = \mathtt{MSE}(\mathbf{y}, \hat{\mathbf{y}})$. During finetuning, the encoder of $\mathcal{P}$ is initialized with the weights from pretaining, and the decoder is initialized randomly. The aspect ratio $H:W$ is set to be $4:3$, with the pretrained positional embedding being interpolated accordingly\cite{kirillov2023segment}. We use lightweight decoders with deconvolution and convolution operations.  

\begin{figure*}
\centering
\includegraphics[width=0.9\linewidth]{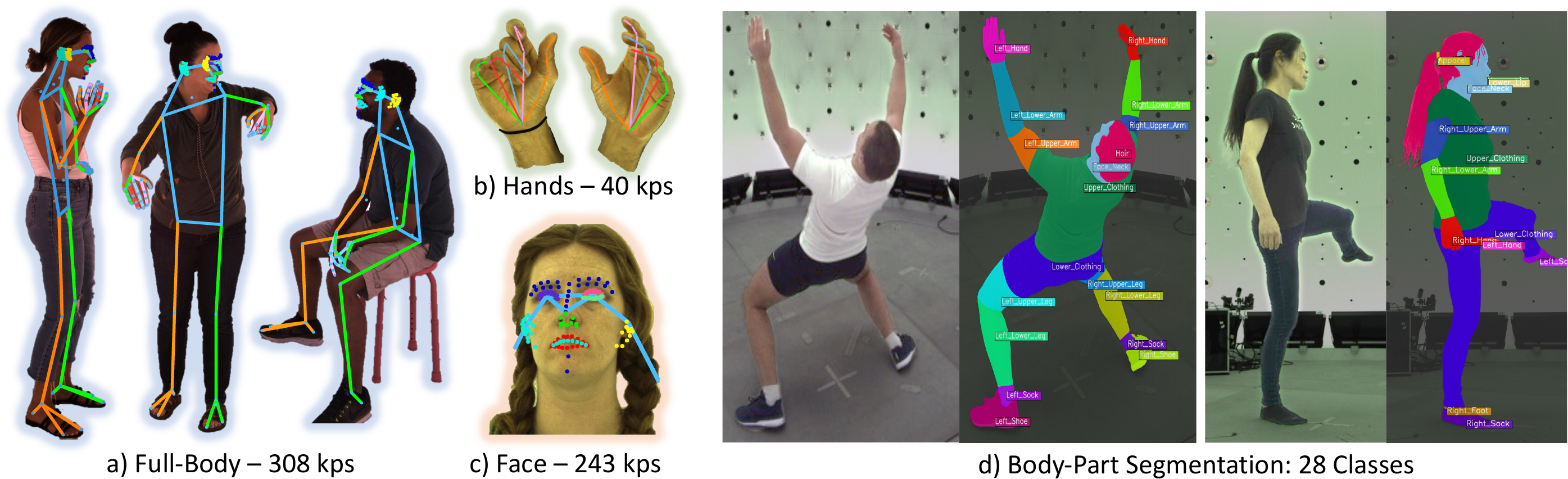}
\caption{Ground-truth annotations for 2D pose estimation and body-part segmentation.}
\vspace*{-0.2in}
\label{figure:pose_seg}
\end{figure*}

We finetune the encoder and decoder in $\mathcal{P}$ across multiple skeletons, including  $K=17$~\cite{lin2014microsoft}, $K=133$~\cite{jin2020whole} and a new highly-detailed skeleton, with $K=308$, as shown in Fig.~\ref{figure:pose_seg} (\textit{Left}). Compared to existing formats with at most $68$ facial keypoints, our annotations consist of $243$ facial keypoints, including representative points around the eyes, lips, nose, and ears. This design is tailored to meticulously capture the nuanced details of facial expressions in the real world.
With these keypoints, we manually annotated $1$ million images at $4K$ resolution from an indoor capture setup. 

\subsection{Body-Part Segmentation}
Commonly referred to as human parsing, body-part segmentation aims to classify pixels in the input image $\mathbf{I}$ into $C$ classes. Most methods~\cite{gong2017look} transform this problem to estimating per-pixel class probabilities to create a probability map $\mathbf{\hat{p}} \in \mathbb{R}^{H \times W \times C}$ such that $\mathbf{\hat{p}} = \mathcal{S}(\mathbf{I})$, where $\mathcal{S}$ is the segmentation model. As outlined previously, we adopt the same encoder-decoder architecture and initialization scheme for $\mathcal{S}$. $\mathcal{S}$ is finetuned to minimize the weighted cross-entropy loss between the actual $\mathbf{p}$ and predicted $\mathbf{\hat{p}}$ probability maps, $\mathcal{L}_\text{seg} = \mathtt{WeightedCE}(\mathbf{p}, \mathbf{\hat{p}})$.

\begin{figure*}[b]
\centering
\includegraphics[width=0.9\linewidth]{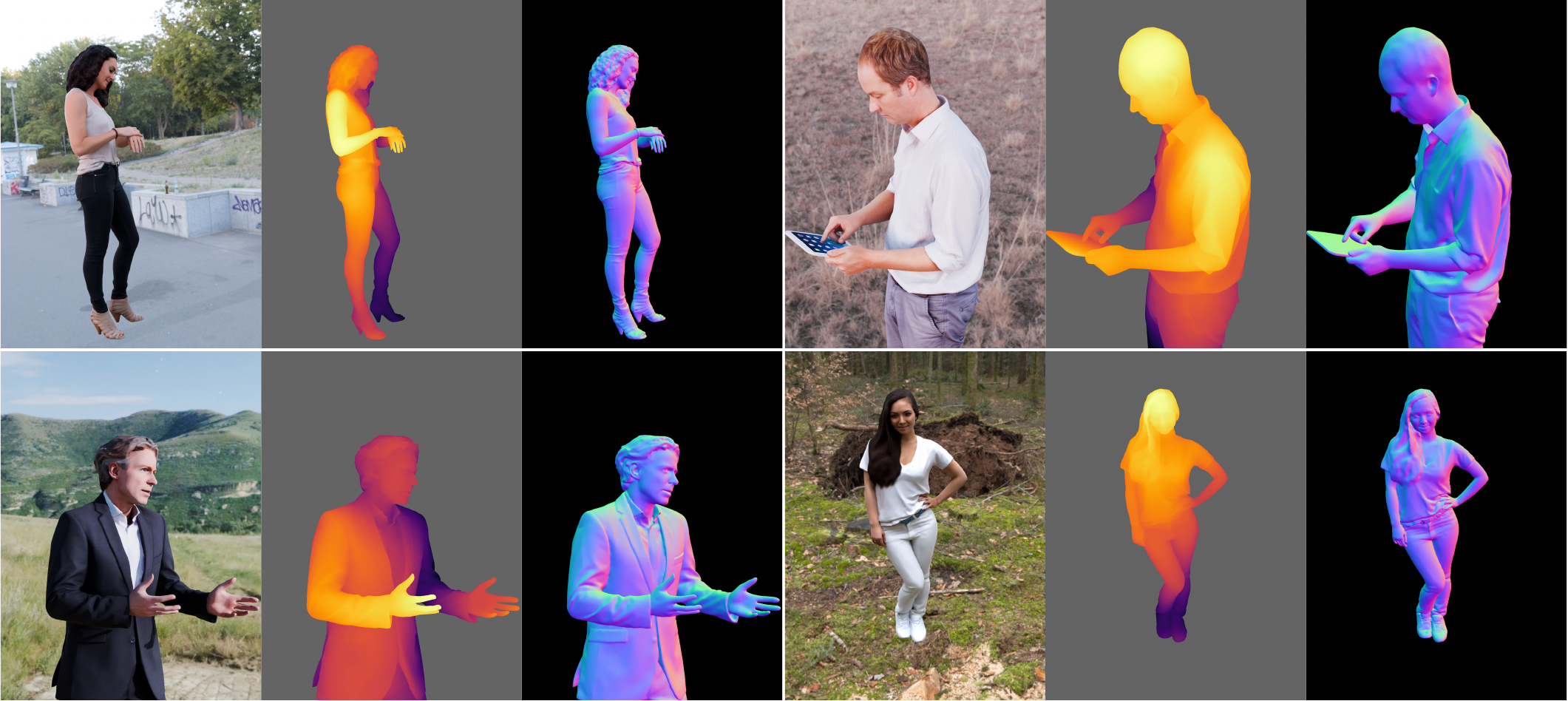}
\caption{Ground-truth synthetic annotations for depth and surface normal estimation.}
\vspace*{-0.2in}
\label{figure:depth_normal}
\end{figure*}

We finetune $\mathcal{S}$ across two part-segmentation vocabularies: a standard set with $C=20$~\cite{gong2017look} and a new larger vocabulary with $C=28$, as illustrated in Fig.\ref{figure:pose_seg} (\textit{Right}). Our proposed vocabulary goes beyond previous datasets in important ways. It distinguishes between the upper and lower halves of limbs and incorporates more detailed classifications such as upper/lower lips, teeth, and tongue. To this end, we manually annotate $100K$ images at $4K$ resolution with this vocabulary.

\subsection{Depth Estimation}
For depth estimation, we adopt the architecture used for segmentation, with the modification that the decoder output channel is set to $1$ for regression. We denote the ground-truth depth map of image $\mathbf{I}$ by $\mathbf{d} \in \mathbb{R}^{H \times W}$, the depth estimator by $\mathcal{D}$, where $\hat{\mathbf{d}} = \mathcal{D}(\mathbf{I})$, and $M$ as the number of human pixels in the image. For the relative depth estimation, we normalize $\mathbf{d}$ to the range $[0, 1]$ using max and min depths in the image. The $\mathcal{L}_\text{depth}$ loss~\cite{eigen2014depth} for $\mathcal{D}$ is defined as follows:
\begin{align}
\Delta \mathbf{d} &= \log(\mathbf{d}) - \log(\hat{\mathbf{d}}), \\
\overline{\Delta \mathbf{d}} &= \frac{1}{M} \sum_{i=1}^{M} \Delta \mathbf{d}_i, \quad & \overline{(\Delta \mathbf{d})^2} &= \frac{1}{M} \sum_{i=1}^{M} (\Delta \mathbf{d}_i)^2, \\
\mathcal{L}_\text{depth} &= \sqrt{\overline{(\Delta \mathbf{d})^2} - \frac{1}{2} (\overline{\Delta \mathbf{d}})^2}.
\end{align}

We render $500,000$ synthetic images using $600$ high-resolution photogrammetry human scans as shown in Fig.~\ref{figure:depth_normal} to obtain a robust monocular depth estimation model with high-fidelity. A random background is selected from a $100$ HDRI environment map collection. We place a virtual camera within the scene, randomly adjusting its focal length, rotation, and translation to capture images and their associated ground-truth depth maps at 4K resolution.
 
\subsection{Surface Normal Estimation}
Similar to previous tasks, we set the decoder output channels of the normal estimator $\mathcal{N}$ to be $3$, corresponding to the $xyz$ components of the normal vector at each pixel. The generated synthetic data is also used as supervision for surface normal estimation. Let $\mathbf{n}$ be the ground-truth normal map for image $\mathbf{I}$ and $\mathbf{\hat{n}} = \mathcal{N}(\mathbf{I})$. Similar to depth, the loss $\mathcal{L}_\text{normal}$ is only computed for human pixels in the image and is defined as follows:
\begin{equation}
\mathcal{L}_\text{normal} = ||\mathbf{n} - \mathbf{\hat{n}}||_1 + (1 - \mathbf{n} \cdot \mathbf{\hat{n}})
\end{equation}

\section{Experiments}
\label{sec:experiments}
\begin{table*}[t]
\centering
\resizebox{4.2in}{!}{
    \setlength{\tabcolsep}{6pt}
    \renewcommand{\arraystretch}{1.1}
    \begin{tabular}{l | cccccc}
    \toprule
    \textbf{Model}    & \textbf{\#Params}  & \textbf{FLOPs} & \textbf{Hidden size} & \textbf{Layers}  & \textbf{Heads} & \textbf{Batch size} \\ 
    \midrule
    Sapiens-0.3B  & 0.336 B     & 1.242 T     & 1024  & 24      & 16      & 98,304       \\
    Sapiens-0.6B  & 0.664 B     & 2.583 T     & 1280  & 32      & 16      & 65,536       \\
    Sapiens-1B    & 1.169 B     & 4.647 T     & 1536  & 40      & 24      & 40,960       \\
    Sapiens-2B    & 2.163 B     & 8.709 T     & 1920  & 48      & 32      & 20,480  \\
    \bottomrule
    \end{tabular}
}
\vspace{-0.1in}
\caption{Sapiens encoder specifications for pretraining on Human-300M dataset.}
\vspace{-0.2in}
\label{table:model_specs}
\end{table*}

In this section, we initially provide an overview of the implementation details. Subsequently, we conduct comprehensive benchmarking across four tasks: pose estimation, part segmentation, depth estimation, and normal estimation.

\begin{figure*}[b]
\centering
\includegraphics[width=0.98\linewidth]{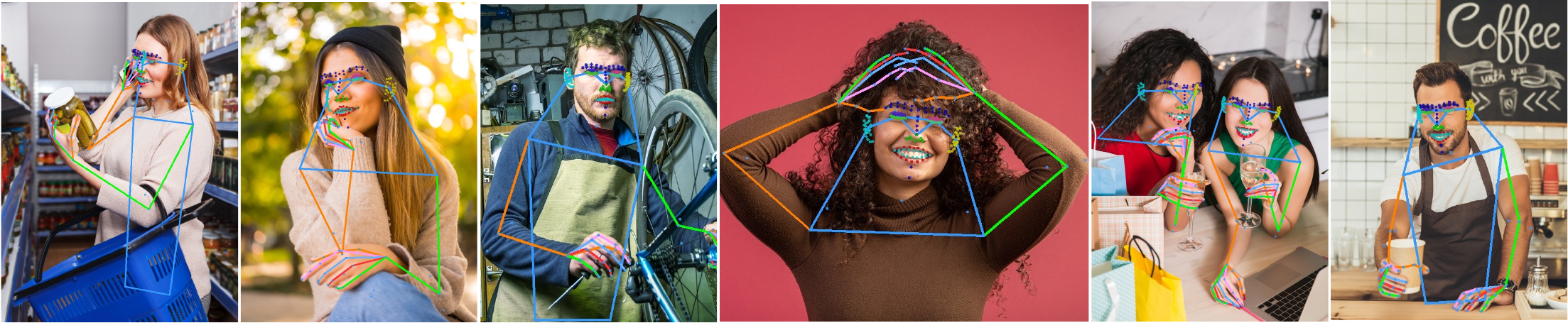}
\vspace*{-0.1in}
\caption{Pose estimation with Sapiens-1B for 308 keypoints on in-the-wild images.}
\vspace*{-0.1in}
\label{figure:experiments_pose}
\end{figure*}

\subsection{Implementation Details}
Our largest model, Sapiens-2B, is pretrained using $1024$ A100 GPUs for $18$ days using PyTorch. 
We use the AdamW~\cite{loshchilov2017decoupled} optimizer for all our experiments. The learning schedule includes a brief linear warm-up, followed by cosine annealing~\cite{loshchilov2016sgdr} for pretraining and linear decay~\cite{lewkowycz2021decay} for finetuning. All models are pretrained from scratch at a resolution of $1024 \times 1024$ with a patch size of $16$. For finetuning, the input image is resized to a 4:3 ratio, \ie $1024 \times 768$. We use standard augmentations like cropping, scaling, flipping, and photometric distortions. A random background from non-human COCO~\cite{lin2014microsoft} images is added for segmentation, depth, and normal prediction tasks. Importantly, we use differential learning rates~\cite{yang2019xlnet} to preserve generalization \ie lower learning rates for initial layers and progressively higher rates for subsequent layers. The layer-wise learning rate decay is set to $0.85$ with a weight decay of $0.1$ for the encoder. We detail the design specifications of Sapiens in Table. \ref{table:model_specs}. Following ~\cite{touvron2023llama2, el2024scalable}, we prioritize scaling models by width rather than depth.
Note that the Sapiens-0.3B model, while architecturally similar to the traditional ViT-Large, consists of twentyfold more FLOPs due to its higher resolution.

\subsection{2D Pose Estimation}

We finetune Sapiens for face, body, feet, and hand ($K=308$) pose estimation on our high-fidelity annotations. For training, we use the \texttt{train} set with $1M$ images and for evaluation, we use the \texttt{test} set, named Humans-5K, with $5K$ images. Our evaluation is top-down~\cite{xu2022vitpose} \ie we use an off-the-shelf detector~\cite{fang2023eva02} for bounding-box and conduct single human pose inference. Table~\ref{table:pose} shows a comparison of our models with existing methods for whole-body pose estimation. We evaluate all methods on $114$ common keypoints between our $308$ keypoint vocabulary and the $133$ keypoint vocabulary from COCO-WholeBody~\cite{jin2020whole}. Sapiens-0.6B surpasses the current state-of-the-art, DWPose-l~\cite{yang2023effective} by $+2.8$ AP. Contrary to DWPose~\cite{yang2023effective}, which utilizes a complex student-teacher framework with feature distillation tailored for the task, Sapiens adopts a general encoder-decoder architecture with large human-centric pretraining.

Interestingly, even with the same parameter count, our models demonstrate superior performance compared to their counterparts. For instance, Sapiens-0.3B exceeds VitPose+-L by $+5.6$ AP, and Sapiens-0.6B outperforms VitPose+-H by $+7.9$ AP. Within the Sapiens family, our results indicate a direct correlation between model size and performance. Sapiens-2B sets a state-of-the-art with $61.1$ AP, a significant improvement of $\mathbf{+7.6}$ \textbf{AP} to the prior art.  Despite fine-tuning with annotations from a indoor capture studio, Sapiens demonstrate robust generalization to real-world, as shown in Fig.~\ref{figure:experiments_pose}.


\begin{table*}[t]
\centering
\resizebox{5.8in}{!}{
\setlength{\tabcolsep}{6pt}
\renewcommand{\arraystretch}{1.2}
\begin{tabular}{l|c|cc|cc|cc|cc|ll}
\toprule
\textbf{Model} & \textbf{Input Size} & \multicolumn{2}{c|}{\textbf{Body}} & \multicolumn{2}{c|}{\textbf{Foot}} & \multicolumn{2}{c|}{\textbf{Face}} & \multicolumn{2}{c|}{\textbf{Hand}} & \multicolumn{2}{c}{\textbf{Whole-body}} \\
& & AP & AR & AP & AR & AP & AR & AP & AR & AP & AR \\
\midrule
DeepPose~\cite{toshev2014deeppose} & $384 \times 288$ & 32.1 & 43.5 & 25.3 & 41.2 & 37.8 & 53.9 & 15.7 & 31.6 & 23.9 & 37.2 \\
SimpleBaseline~\cite{xiao2018simple} & $384 \times 288$ & 52.3 & 60.1 & 49.8 & 62.5 & 59.6 & 67.3 & 41.4 & 51.8 & 44.6 & 53.7 \\
HRNet~\cite{sun2019deep} & $384 \times 288$ & 55.8 & 62.6 & 45.2 & 55.4 & 58.9 & 64.5 & 39.3 & 47.6 & 45.7 & 53.9 \\
ZoomNAS~\cite{xu2022zoomnas} & $384 \times 288$ & 59.7 & 66.3 & 48.1 & 57.9 & 74.5 & 79.2 & 49.8 & 60.6 & 52.1 & 60.7 \\
ViTPose+-L~\cite{xu2022vitpose+} & $256 \times 192$ & 61.0 & 66.8 & 62.4 & 68.2 & 50.1 & 55.7 & 41.5 & 47.3 & 47.8 & 53.6 \\
ViTPose+-H~\cite{xu2022vitpose+} & $256 \times 192$ & 61.6 & 67.4 & 63.2 & 69.0 & 50.7 & 56.3 & 42.0 & 47.8 & 48.3 & 54.1 \\
RTMPose-x~\cite{jiang2023rtmpose} & $384 \times 288$ & 57.1 & 63.7 & 55.3 & 66.8 & 74.4 & 78.5 & 46.3 & 55.0 & 51.9 & 59.6 \\
DWPose-m~\cite{yang2023effective} & $256 \times 192$ & 54.2 & 61.4 & 49.9 & 63.0 & 68.5 & 74.2 & 40.1 & 50.0 & 47.7 & 55.8 \\
DWPose-l~\cite{yang2023effective} & $384 \times 288$ & 57.9 & 64.2 & 56.5 & 67.4 & 74.3 & 78.4 & 49.3 & 57.4 & 53.1 & 60.6 \\
\midrule
Sapiens-0.3B (Ours) & $1024 \times 768$ & 58.1 & 64.5 & 56.8 & 67.7 & 74.5 & 78.6 & 49.6 & 57.7 & 53.4 \scriptsize{(+0.3)} & 60.9 \scriptsize{(+0.3)} \\
Sapiens-0.6B (Ours) & $1024 \times 768$ & 59.8 & 65.5 & 64.7 & 72.3 & 75.2 & 79.0 & 52.1 & 60.3 & 56.2 \scriptsize{(+2.8)} & 62.4 \scriptsize{(+2.1)} \\
Sapiens-1B (Ours) & $1024 \times 768$ & 62.9 & 68.2 & 68.3 & 75.1 & 76.4 & 79.7 & 55.9 & 63.4 & 59.4 \scriptsize{(+5.9)} & 65.3 \scriptsize{(+5.1)} \\
Sapiens-2B (Ours) & $1024 \times 768$ & \textbf{64.7} & \textbf{69.9} & \textbf{69.4} & \textbf{76.2} & \textbf{76.9} & \textbf{79.9} & \textbf{57.1} & \textbf{64.4} & \textbf{61.1\scriptsize{(+7.6)}} & \textbf{67.1\scriptsize{(+7.0)}} \\
\bottomrule
\end{tabular}
}
\vspace{-0.1in}
\caption{Pose estimation results on Humans-5K \texttt{test} set. Flip test is used.}
\vspace{-0.2in}
\label{table:pose}
\end{table*}

\subsection{Body-Part Segmentation}

\begin{figure*}[b]
\centering
\includegraphics[width=\linewidth]{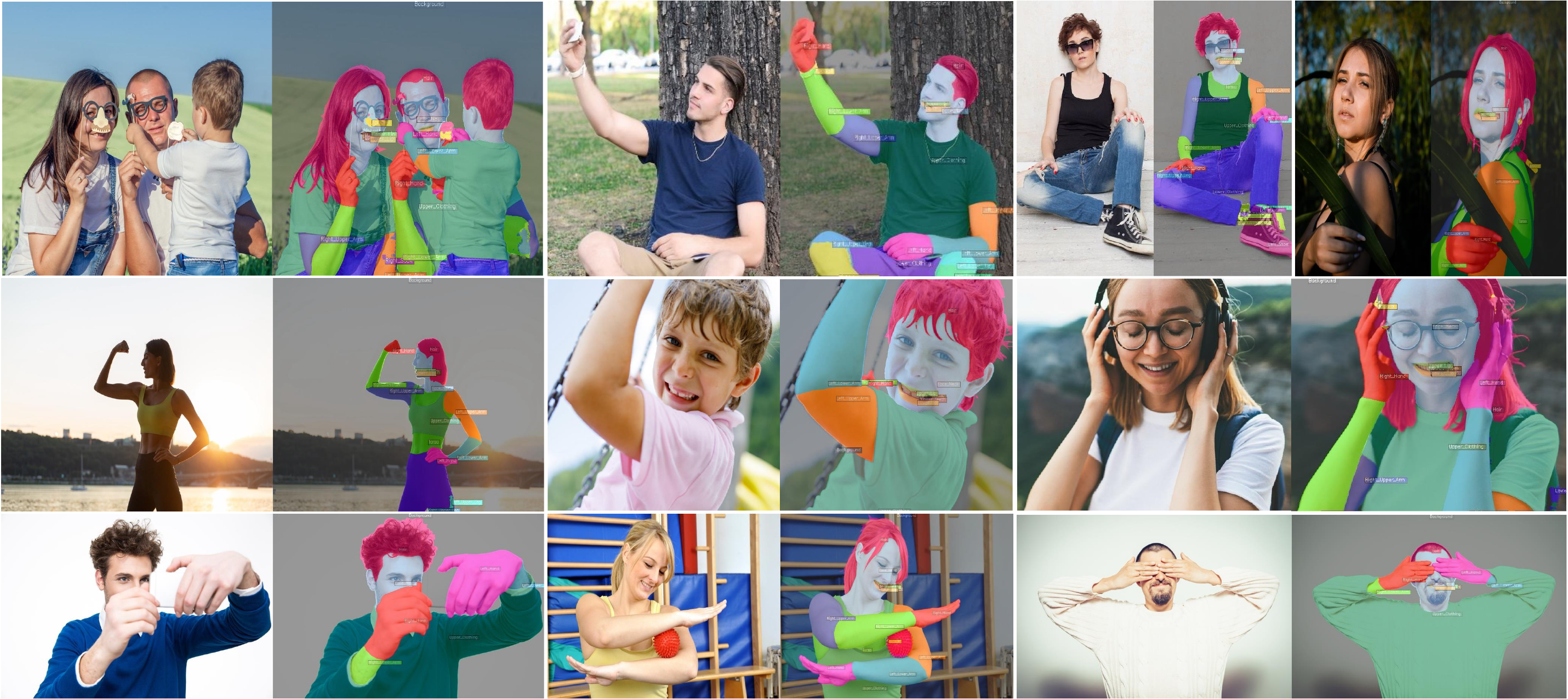}
\vspace{-0.2in}
\caption{Body-part segmentation with Sapiens-1B for 28 categories on single and multi-human images.}
\label{figure:experiments_seg}
\end{figure*}

\vspace{-0.1in}
We fine-tune and evaluate our annotations with a segmentation vocabulary of $28$ classes. Our \texttt{train} set consists of $100K$ images, and the \texttt{test} set, Humans-2K, consists of $2K$ images. We compare Sapiens with existing body-part segmentation methods fine-tuned on our \texttt{train} set. Importantly, we use suggested pretrained checkpoints by each method as initialization. Similar to pose, we observe generalization to segmentation as shown in Table \ref{table:seg}. 

\begin{table}[b]
\centering
\vspace{-0.3in}
\resizebox{2.2in}{!}{
    \setlength{\tabcolsep}{6pt}
    \renewcommand{\arraystretch}{1.1}
    \begin{tabular}{lcc}
    \toprule
    \textbf{Model} & \textbf{mIoU(\%)} & \textbf{mAcc(\%)} \\
    \midrule
    FCN*~\cite{long2014fully} & 48.2 & 57.6 \\
    SegFormer*~\cite{xie2021segformer} & 53.5 & 62.9  \\
    Mask2Former*~\cite{cheng2021masked} & 58.7 & 68.3 \\
    DeepLabV3+*~\cite{chen2018encoder} & 64.1 & 74.8 \\
    \midrule
    Sapiens-0.3B (Ours) & $76.7$ & $86.1$ \\
    Sapiens-0.6B (Ours) & $77.8$ & $86.3$ \\
    Sapiens-1B (Ours) & $79.9$ & $89.1$ \\
    Sapiens-2B (Ours) & $\textbf{81.2}$ & $\textbf{89.4}$ \\
    \bottomrule
    \end{tabular}
}
\vspace{-0.1in}
\caption{We report mIoU and mAcc on Humans-2K \texttt{test} set. Methods with * are trained by us.}
\vspace{-0.1in}
\label{table:seg}
\end{table}

Interestingly, our smallest model, Sapiens-0.3B outperforms existing state-of-the-art segmentation methods like Mask2Former~\cite{cheng2021masked} and DeepLabV3+~\cite{chen2018encoder} by $12.6$ mIoU due to its higher resolution and large human-centric pretraining. Furthermore, increasing the model size improves segmentation performance. Sapiens-2B achieves the best performance of $81.2$ mIoU and $89.4$ mAcc on the \texttt{test} set. Fig.~\ref{figure:experiments_seg} shows the qualitative results of our models.

\newpage
\clearpage
\subsection{Depth Estimation}
\vspace{-0.1in}
\begin{table*}[t]
\centering
\setlength{\tabcolsep}{6pt}
\renewcommand{\arraystretch}{1.1}
\resizebox{6in}{!}{
\begin{tabular}{l|ccc|ccc|ccc|ccc}
\toprule
\textbf{Method} & \multicolumn{3}{c|}{\textbf{TH2.0-Face} } & \multicolumn{3}{c|}{\textbf{TH2.0-UprBody} } & \multicolumn{3}{c|}{\textbf{TH2.0-FullBody} } & \multicolumn{3}{c}{\textbf{Hi4D} } \\
                & RMSE $\downarrow$ & AbsRel $\downarrow$ & $\delta_1$ $\uparrow$ & RMSE & AbsRel & $\delta_1$ & RMSE & AbsRel & $\delta_1$ & RMSE & AbsRel & $\delta_1$ \\
\midrule
MiDaS-L~\cite{birkl2023midas}    & 0.114 & 0.097 & 0.925 & 0.398 & 0.271 & 0.868 & 0.701 & 0.689 & 0.782 & 0.261 & 0.082 & 0.975 \\
MiDaS-Swin2~\cite{birkl2023midas}    & 0.050 & 0.036 & 0.995 & 0.122 & 0.081 & 0.948 & 0.292 & 0.171 & 0.862 & 0.209 & 0.063 & 0.997 \\
DepthAny-B~\cite{yang2024depth}    & 0.039 & 0.026 & 0.999 & 0.048 & 0.028 & 0.999 & 0.061 & 0.030 & 0.999 & 0.143 & \textbf{0.034} & 0.997 \\
DepthAny-L~\cite{yang2024depth}    & 0.039 & 0.027 & 0.999 & 0.048 & 0.027 & 0.999 & 0.060 & 0.030 & 0.999 & 0.147 & 0.035 & 0.997 \\
\midrule
Sapiens-0.3B (Ours)   & 0.012 & 0.008 & 1.000 & 0.015 & 0.009 & 1.000 & 0.021 & 0.010 & 1.000 & 0.148 & 0.046 & 1.000 \\
Sapiens-0.6B (Ours)   & 0.011 & 0.008 & 1.000 & 0.015 & 0.009 & 1.000 & 0.021 & 0.010 & 1.000 & 0.142 & 0.044 & 1.000 \\
Sapiens-1B (Ours)   & 0.009 & 0.006 & 1.000 & 0.012 & 0.007 & 1.000 & 0.019 & 0.009 & 1.000 & 0.125 & 0.039 & 1.000 \\
Sapiens-2B (Ours)   & \textbf{0.008} & \textbf{0.005} & \textbf{1.000} & \textbf{0.010} & \textbf{0.006} & \textbf{1.000} & \textbf{0.016} & \textbf{0.008} & \textbf{1.000} & \textbf{0.114} & 0.036 & \textbf{1.000} \\
\bottomrule
\end{tabular}
}
\vspace{-0.1in}
\caption{Comparison of Sapiens for monocular depth estimation on human images.}
\vspace{-0.2in}
\label{table:depth}
\end{table*}

\begin{figure*}[b]
\centering
\vspace*{-0.1in}
\includegraphics[width=0.95\linewidth]{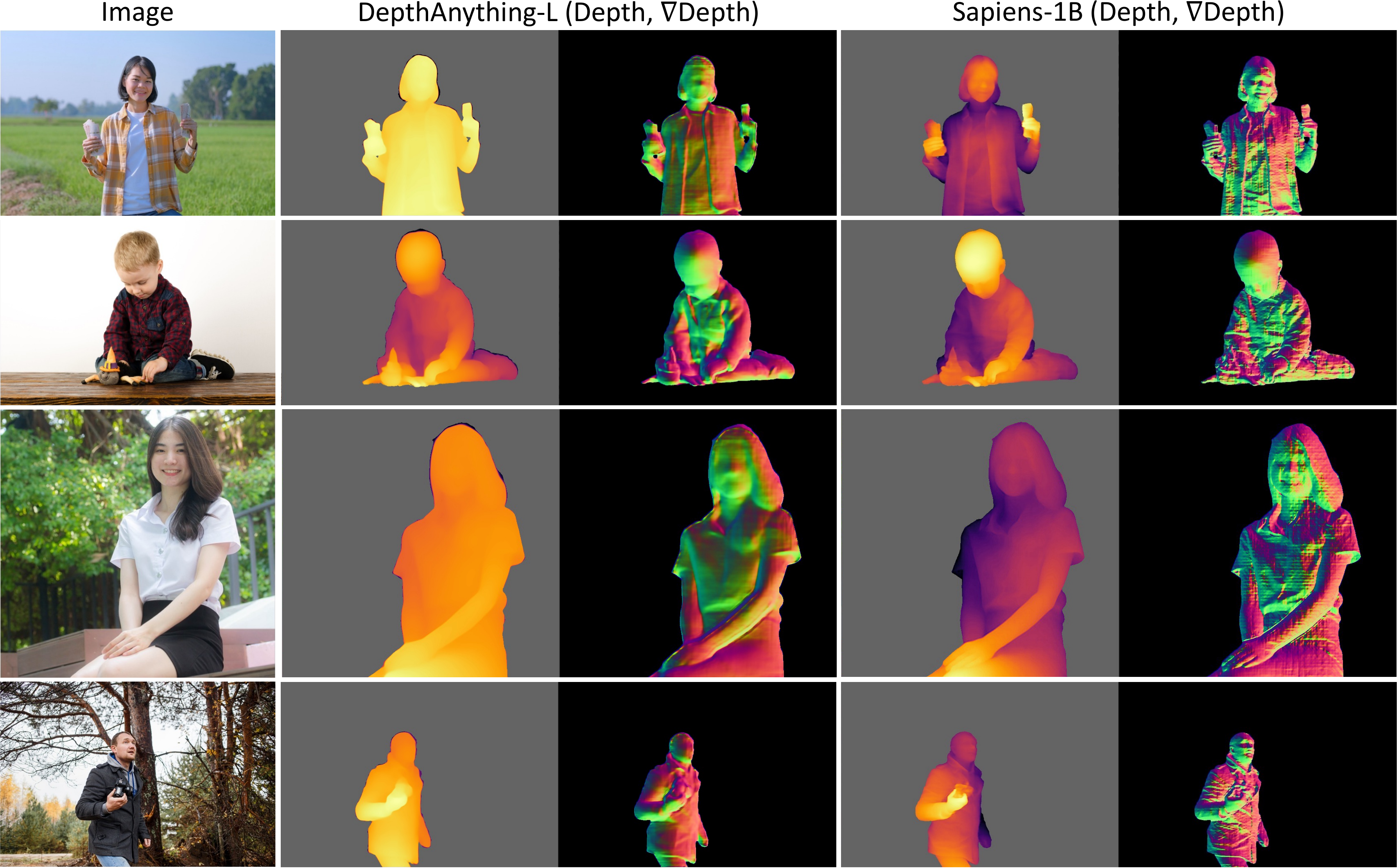}
\caption{We compare our depth prediction with DepthAnything~\cite{yang2024depth}. To showcase the consistency of predicted depth, we also visualize the $\nabla$depth as pseudo surface normals.}
\vspace*{-0.3in}
\label{figure:experiments_depth}
\end{figure*}

We evaluate our models on THuman2.0\cite{yu2021function4d} and Hi4D\cite{yin2023hi4d} datasets for depth estimation. THuman2.0 consists of $526$ high-quality human scans, from which we derive three sets of images for testing: a) face, b) upper body, and c) full body using a virtual camera. THuman2.0 with $1578$ images thus enables the evaluation of our models' performance on single-human images across multiple scales. Conversely, the Hi4D dataset focuses on multi-human scenarios, with each sequence showcasing two subjects engaged in activities involving human-human interactions. We select sequences from pair 28, 32, and 37, featuring 6 unique subjects from camera 4, totaling $1195$ multi-human real images for testing. We follow the relative-depth evaluation protocols established by MiDaS-v3.1~\cite{birkl2023midas}, reporting standard metrics such as AbsRel and $\delta_1$. In addition, we also report RMSE as our primary metric since $\delta_1$ does not effectively reflect performance in human scenes characterized by subtle depth variations.

Table \ref{table:depth} compares our models with existing state-of-the-art monocular depth estimators. Sapiens-2B, finetuned solely on synthetic data, remarkably outperforms prior art across all single-human scales and multi-human scenarios. We observe a $20\%$ RMSE reduction compared to the top-performing Depth-Anything model on Hi4D images. It is important to highlight that while baseline models are trained on a variety of scenes, Sapiens specializes in human-centric depth estimation. Fig. \ref{figure:experiments_depth} presents a qualitative comparison of depth estimation between Sapiens-1B and DepthAnything-L. To ensure a fair comparison, the predicted depth is renormalized using the human mask in the baseline visualizations.

\newpage
\clearpage
\subsection{Surface Normal Estimation}
\vspace{-0.1in}
\begin{table*}[t]
\centering
\vspace*{-0.2in}
\setlength{\tabcolsep}{6pt}
\renewcommand{\arraystretch}{1.2}
\resizebox{5.4in}{!}{
\begin{tabular}{l|cc|ccc|cc|ccc}
\toprule
\multirow{3}{*}{\textbf{Method}} & \multicolumn{5}{c|}{\textbf{THuman2.0}~\cite{yu2021function4d}}  & \multicolumn{5}{c}{\textbf{Hi4D}~\cite{yin2023hi4d}} \\

& \multicolumn{2}{c|}{$\text{Angular Error}^\circ$} & \multicolumn{3}{c|}{\% Within $t^\circ$}   & \multicolumn{2}{c|}{$\text{Angular Error}^\circ$} & \multicolumn{3}{c}{\% Within $t^\circ$} \\
                & Mean  & Median & $11.25^\circ$  & $22.5^\circ$ & $30^\circ$ & Mean  & Median & $11.25^\circ$  & $22.5^\circ$ & $30^\circ$\\
\midrule
PIFuHD~\cite{saito2020pifuhd}  & 30.51 & 27.13 & 15.81 & 42.97 & 58.86 & 22.39 & 19.26 & 22.98 & 60.14 & 77.02 \\
HDNet~\cite{jafarian2021learning}  & 34.82 & 30.60 & 17.44 & 39.26 & 54.51 & 28.60 & 26.85 & 19.08 & 57.93 & 70.14 \\
ICON~\cite{xiu2022icon}  & 28.74 & 25.52 & 22.81 & 47.83 & 63.73 & 20.18 & 17.52 & 26.81 & 66.34 & 82.73 \\
ECON~\cite{xiu2022econ}  & 25.45 & 23.67 & 32.95 & 55.86 & 69.03 & 18.46 & 16.47 & 29.35 & 68.12 & 84.88 \\

\midrule
Sapiens-0.3B & 13.02 & 10.33 & 57.37 & 86.20 & 92.7 & 15.04 & 12.22 & 47.07 & 81.49 & 90.70 \\
Sapiens-0.6B & 12.86 & 10.23 & 57.85 & 86.68 & 93.30 & 14.06 & 11.47 & 50.59 & 84.37 & 92.54 \\
Sapiens-1B & 12.11 & 9.40 & 61.97 & 88.03 & 93.84 & 12.18 & \textbf{9.59} & \textbf{60.36} & 88.62 & 94.44 \\
Sapiens-2B & \textbf{11.84} & \textbf{9.16} & \textbf{63.16} & \textbf{88.60} & \textbf{94.18} & \textbf{12.14} & 9.62 & 60.22 & \textbf{89.08} & \textbf{94.74} \\

\bottomrule
\end{tabular}
}
\vspace{-0.1in}
\caption{Comparison of Sapiens for surface normal estimation on human images.}
\vspace{-0.2in}
\label{table:normal}
\end{table*}

\begin{figure*}[b]
\centering
\vspace*{-0.1in}
\includegraphics[width=0.7\linewidth]{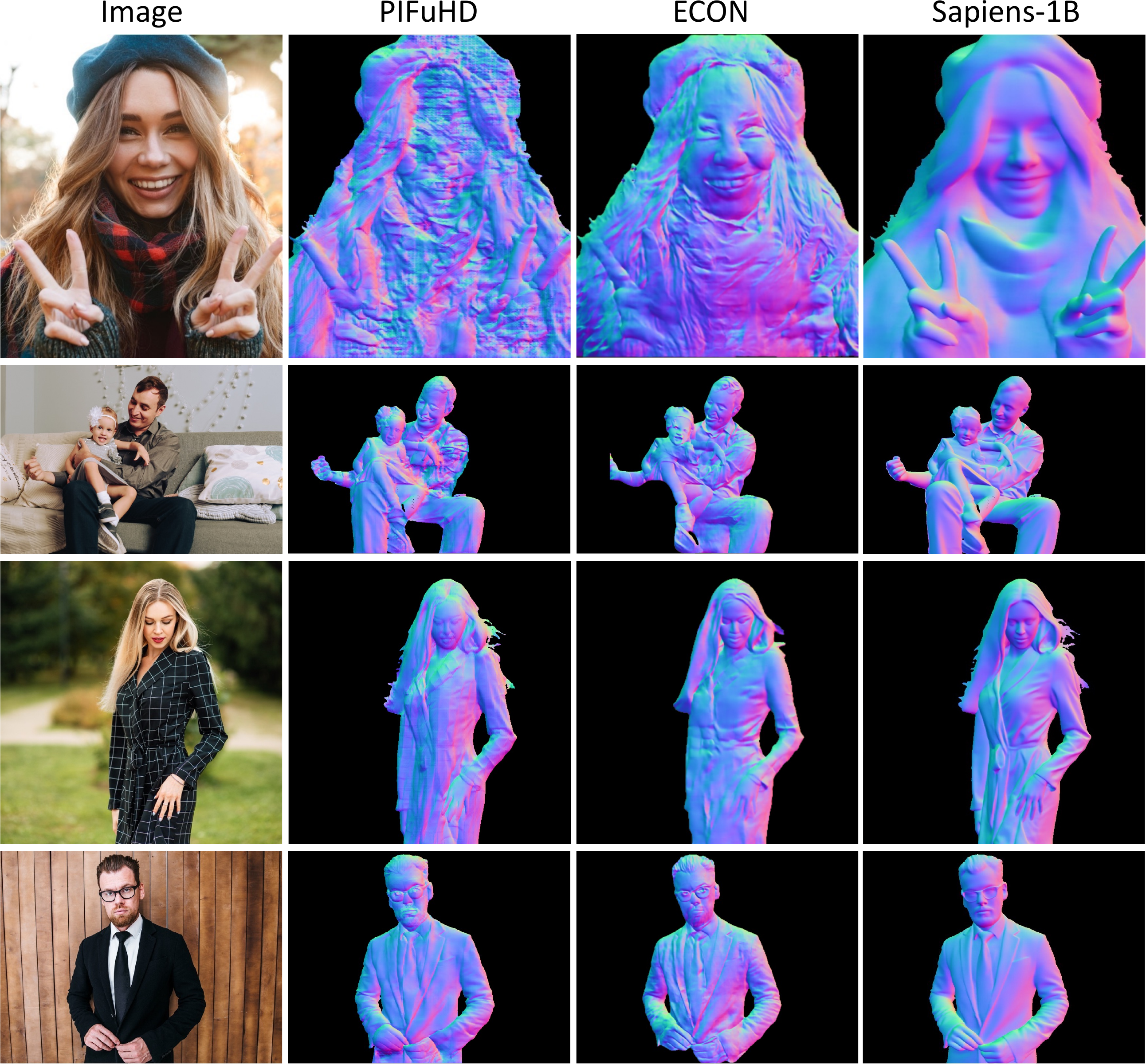}
\vspace*{-0.1in}
\caption{Qualitative comparison of Sapiens-1B with PIFuHD~\cite{saito2020pifuhd} and ECON~\cite{xiu2022econ} for surface normal estimation on in-the-wild images.}
\vspace*{-0.2in}
\label{figure:experiments_normal}
\end{figure*}

The datasets for surface normal evaluation are identical to those used for depth estimation. Following~\cite{eftekhar2021omnidata}, we report the mean and median angular error, along with the percentage of pixels within $t^\circ$ error for $t \in \{11.25^\circ, 22.5^\circ, 30^\circ\}$. Table \ref{table:normal} compares our models with existing human-specific surface normal estimators. All our models outperform existing methods by a significant margin. Sapiens-2B achieves a mean error of around $12^\circ$ on the THuman2.0 (single-human) and Hi4D (multi-human) datasets. We qualitatively compare Sapiens-1B with PIFuHD~\cite{saito2020pifuhd} and ECON~\cite{xiu2022econ} for surface normal estimation in Figure \ref{figure:experiments_normal}. Note that PIFuHD~\cite{saito2020pifuhd} is trained with the identical set of 3D scans as ours, and ECON~\cite{xiu2022econ} is trained with 4000 scans that are a super set of our 3D scan data.

\subsection{Discussion}
\noindent
\textbf{Importance of Pretraining Data Source.} The feature quality is closely linked to the pretraining data quality. We assess the importance of pretraining on various data sources for human-centric tasks by pretraining Sapiens-0.3B on each dataset under identical training schedules and number of iterations. 
We fine-tune the model on each task and select early checkpoints for evaluation, reasoning that early-stage fine-tuning better reflects the model's generalization capability. We investigate the impact of pretraining at scale on general images (which may include humans) versus exclusively human images using Sapiens. We randomly select $100$ million and $300$ million general images from our $1$ billion image corpus to create the General-100M and General-300M datasets, respectively. Table~\ref{table:pretrain} showcases the comparison of pretraining outcomes. We report mAP for pose on Humans-5K, mIoU for segmentation on Humans-2K, RMSE for depth on THuman2.0, and mean angular error in degrees for normal estimation on Hi4D. Aligned with findings from~\cite{xu2022vitpose+}, our results show that pretraining with Human300M leads to superior performance across all metrics, highlighting the benefits of human-centric pretraining within a fixed computational budget. 

\begin{figure}
\centering
\vspace*{-0.2in}
\includegraphics[width=0.85\linewidth]{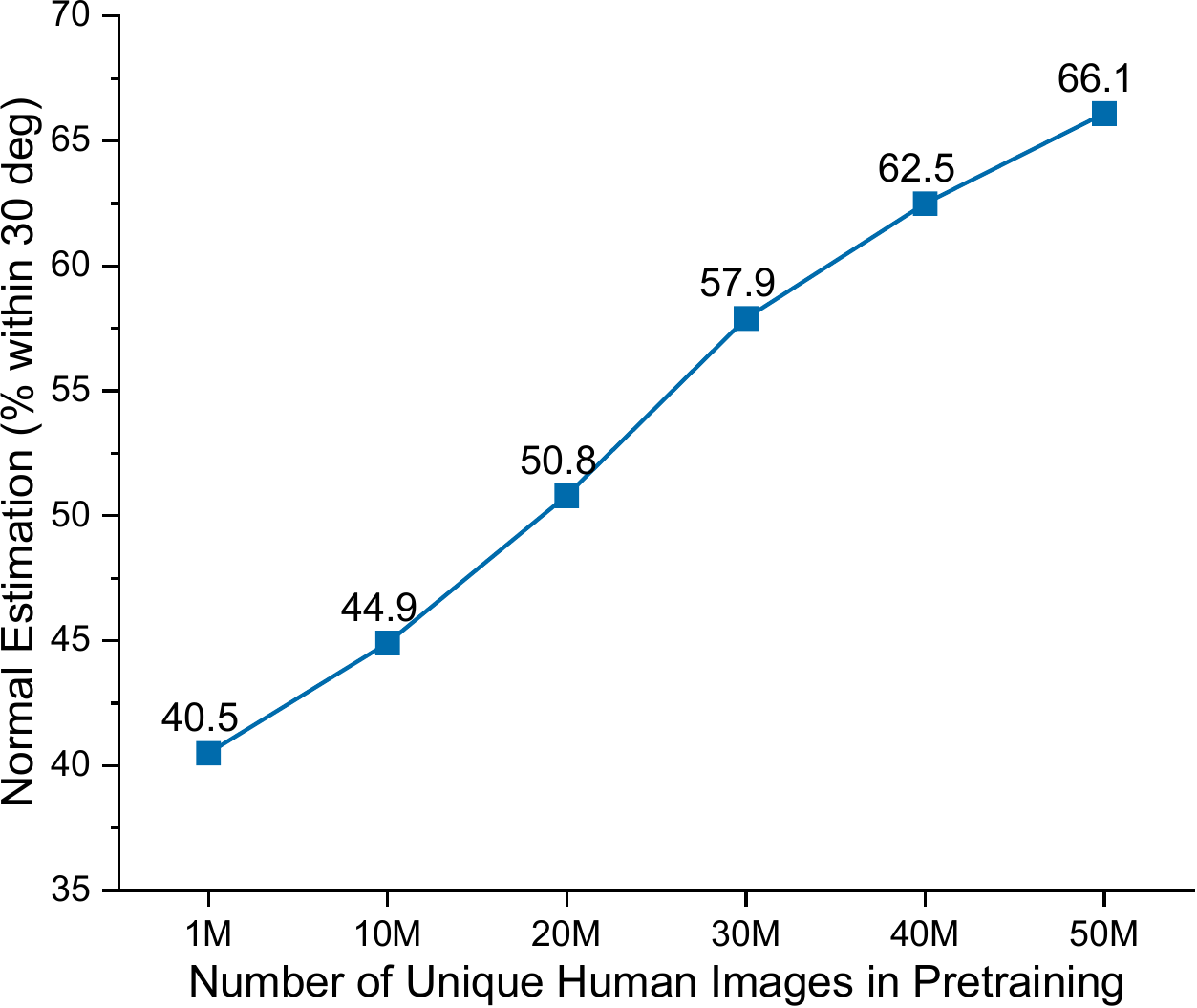}
\caption{Sapiens-0.3B's normal estimation performance with unique human images seen during pretraining.}
\label{figure:discussion_data_size}
\vspace*{-0.2in}
\end{figure}

\begin{figure*}[b]
\centering
\includegraphics[width=0.85\linewidth]{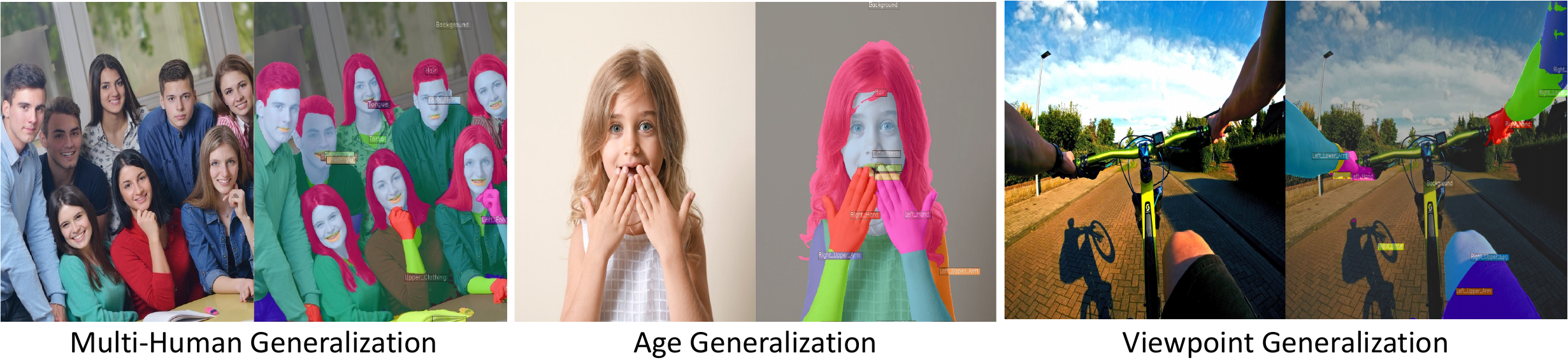}
\vspace*{-0.1in}
\caption{\textit{Sapiens} achieve broad generalization via large human-centric pretraining.}
\vspace*{-0.2in}
\label{figure:discussion_generalization}
\end{figure*}

\begin{table}[b]
\centering
\vspace{-0.2in}
\resizebox{3.4in}{!}{
    \setlength{\tabcolsep}{4pt}
    \renewcommand{\arraystretch}{1.2}
    
    \begin{tabular}{l|c|cccc}
    \toprule
    \textbf{Pretraining Source} & \textbf{\#Images} & \textbf{Pose} ($\uparrow$) &  \textbf{Seg}($\uparrow$) &  \textbf{Depth}($\downarrow$)  & \textbf{Normal}($\downarrow$)  \\
\midrule
Random Initialization & - & 30.2 & 40.3 & 0.720 & 35.4\\
General-100M & 100M & 35.7 & 50.1 & 0.351 & 27.5 \\
General-300M & 300M & 37.3 & 52.8 & 0.347 & 26.8 \\
Humans-100M & 100M & 43.6 & 61.2 & 0.316 & 24.0 \\
Humans-300M (Full) & 300M & \textbf{47.0} & \textbf{66.5} & \textbf{0.288} & \textbf{21.8}\\
    \bottomrule
    \end{tabular}
}
\vspace{-0.1in}
\caption{Comparison of Sapiens-0.3B pretrained on various data sources. A domain-specific pretraining yields superior results compared to general data sources.}
\vspace{-0.2in}
\label{table:pretrain}
\end{table}

We also study the effect of number of unique human images seen during pretraining with normal estimation performance. We report $\%$ within $30^\circ$. Again, we maintain identical conditions for Sapiens-0.3B pretraining and finetuning. Fig.\ref{figure:discussion_data_size} shows a steady improvement in performance as the pretraining data size increases without saturation. In summary, the diversity of human images observed during pretraining directly correlates with improved generalization to down-stream tasks.

\vspace{2mm} \noindent 
\textbf{Zero-Shot Generalization.} Our models exhibit broad generalization to a variety of settings. For instance, in segmentation, Sapiens are finetuned on single-human images with limited subject diversity, minimal background variation, and solely third-person views (see Fig.~\ref{figure:pose_seg}).  Nevertheless, our large-scale pretraining enables generalization across number of subjects, varying ages, and egocentric views, as shown in Fig.~\ref{figure:discussion_generalization}. These observations similarly hold for other tasks.



\vspace{2mm} \noindent 
\textbf{Limitations.} While our models generally perform well, they are not perfect. Human images with complex/rare poses, crowding, and severe occlusion are challenging (see supplemental for details). Although aggressive data augmentation and a detect-and-crop strategy could mitigate these issues, we envision our models as a tool for acquiring large-scale, real-world supervision with human-in-the-loop to develop the next generations of human vision models.

\section{Conclusion}
\label{sec:conclusion}
\textit{Sapiens} represents a significant step toward elevating human-centric vision models into the realm of foundation models. Our models demonstrate strong generalization capabilities on a variety of human-centric tasks. We attribute the state-of-the-art performance of our models to: 
(i) large-scale pretraining on a large 
curated dataset, which is specifically tailored to understanding humans,
(ii) scaled high-resolution and high-capacity vision transformer backbones, and
(iii) high-quality annotations on augmented studio and synthetic data.
We believe that these models can become a key building block for a multitude of downstream tasks,
and provide access to high-quality vision backbones to a significantly wider part of the community. 
A potential direction for future work would be extending \textit{Sapiens} to 3D and multi-modal datasets.

\vspace{2mm} \noindent 
\textbf{Acknowledgements:} We would like to acknowledge He Wen and Srivathsan Govindarajan for their contributions with training, and optimizing Sapiens.

{
    \small
    \bibliographystyle{ieeenat_fullname}
    \bibliography{main}
}

\end{document}